\documentclass[11pt]{article}

\usepackage[final]{acl}

\usepackage{times}
\usepackage{latexsym}

\usepackage[T1]{fontenc}

\usepackage[utf8]{inputenc}
\usepackage{microtype}

\usepackage{inconsolata}

\usepackage{graphicx}
\usepackage{amsmath}
\usepackage{array}
\usepackage{colortbl}
\usepackage{xcolor}
\usepackage{makecell}
\usepackage{amssymb}  
\usepackage{multirow}
\usepackage{tabularx}
\usepackage{tcolorbox}
\usepackage{xcolor}
\usepackage{graphicx}
\usepackage{courier}
\usepackage{CJKutf8}

\newcommand{\jp}[1]{%
  {\begingroup
   \normalfont
   \catcode`\ =10 
   \begin{CJK*}{UTF8}{ipxm}%
   #1%
   \end{CJK*}%
   \endgroup}%
}

\title{FairQE: Multi-Agent Framework for Mitigating Gender Bias in Translation Quality Estimation}

  \author{
  Jinhee Jang$^{1}$ \enspace
  Juhwan Choi$^{2}$\footnotemark[2]\enspace
  Dongjin Lee$^{1}$\footnotemark[2]\enspace
  Seunguk Yu$^{1}$ \enspace
  Youngbin Kim$^{1}$ \\
  $^1$Chung-Ang University \\
  $^2$AITRICS \\
  \texttt{\{jinheejang, dongjinlee30, bokju128, ybkim85\}@cau.ac.kr}, \\
  \texttt{jhchoi@aitrics.com}
}

\begin{document}
\maketitle

\footnotetext[2]{Equal contribution.}

\begin{abstract}
Quality Estimation (QE) aims to assess machine translation quality without reference translations, but recent studies have shown that existing QE models exhibit systematic gender bias. In particular, they tend to favor masculine realizations in gender-ambiguous contexts and may assign higher scores to gender-misaligned translations even when gender is explicitly specified.
To address these issues, we propose FairQE, a multi-agent-based, fairness-aware QE framework that mitigates gender bias in both gender-ambiguous and gender-explicit scenarios. FairQE detects gender cues, generates gender-flipped translation variants, and combines conventional QE scores with LLM-based bias-mitigating reasoning through a dynamic bias-aware aggregation mechanism. This design preserves the strengths of existing QE models while calibrating their gender-related biases in a plug-and-play manner.
Extensive experiments across multiple gender bias evaluation settings demonstrate that FairQE consistently improves gender fairness over strong QE baselines. Moreover, under MQM-based meta-evaluation following the WMT 2023 Metrics Shared Task, FairQE achieves competitive or improved general QE performance. These results show that gender bias in QE can be effectively mitigated without sacrificing evaluation accuracy, enabling fairer and more reliable translation evaluation.
\end{abstract}

\section{Introduction}
\label{sec:intro}
Quality Estimation (QE) aims to automatically assess the quality of machine translation (MT) outputs without relying on human-written reference translations \cite{zhao2024handcrafted}. By removing the dependency on references, QE offers a practical and scalable alternative for translation evaluation, which has led to growing interest from both the MT and evaluation communities \cite{rei2022cometkiwi, mehandru2023physician, lavie2025findings}.

\begin{figure}[t]
    \centering
    \includegraphics[width=\columnwidth]{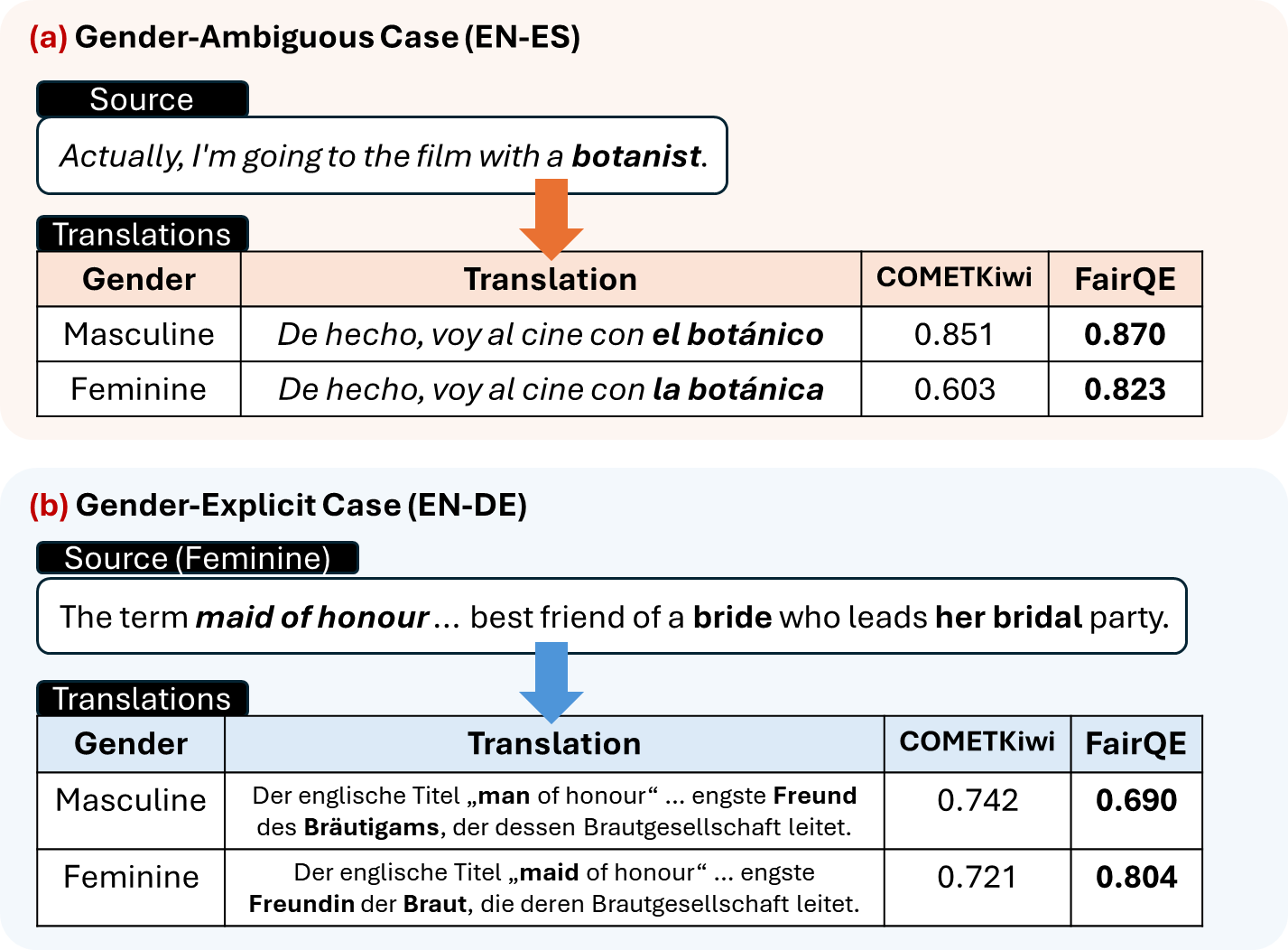}
    \caption{
    Illustration of gender bias in QE model \cite{rei2022cometkiwi}. (a) In gender-ambiguous cases, masculine translations receive higher QE scores despite the absence of gender cues in the source. (b) In gender-explicit cases requiring feminine forms, QE models may still assign higher scores to masculine translations. Our proposed FairQE aims to alleviate such gender biases with a fairer evaluation.}
    \label{fig:overview}
\end{figure}

Despite their effectiveness, recent studies have shown that existing QE models may exhibit systematic biases in gender-related contexts \cite{savoldi2021gender, filandrianos2025gambit, mastromichalakis2025assumed}. In particular, two scenarios have been identified as especially problematic \cite{zaranis2025watching}. First, in gender-ambiguous cases, where the source sentence does not explicitly specify gender, QE models often assign higher scores to translations that realize a specific gender, typically masculine forms, as illustrated in Figure~\ref{fig:overview}(a). Second, in gender-explicit cases, where the source sentence clearly requires a feminine form, QE models may still favor masculine translations. This leads to a phenomenon known as preference inversion, shown in Figure~\ref{fig:overview}(b).

These biases undermine the fairness of QE-based evaluation and may have cascading effects on downstream decision-making processes that rely on QE scores, such as model selection, data filtering, and deployment monitoring \cite{peter2023there}. In particular, reinforcing gender preferences in gender-ambiguous contexts runs counter to broader societal efforts to discourage the reproduction of unnecessary gender stereotypes and to promote inclusive language use \cite{sun2019mitigating, stanczak2021survey}. While a growing body of work has focused on identifying and analyzing gender bias in QE models \cite{savoldi2024harm, zaranis2025watching, filandrianos2025gambit}, relatively few approaches have proposed concrete methodological solutions to mitigate such biases \cite{behnke2022bias, huang2023improving,lee2024gendercontrol}.

To address these limitations, we propose FairQE, a multi-agent-based QE framework designed to mitigate gender-related biases. FairQE jointly addresses both gender-ambiguous and gender-explicit scenarios to provide more equitable translation quality estimates. Specifically, FairQE first detects gender cues to distinguish between ambiguous and explicit cases, and then generates gender-flipped translation variants for each instance. During evaluation, it combines quantitative scores from conventional QE models with large language model (LLM)-based bias-mitigation reasoning, dynamically aggregating these signals according to the estimated severity of gender bias. This design preserves the strengths of existing QE models while effectively calibrating their potential gender bias. It is also model-agnostic, which enables broad applicability across different QE architectures.

We empirically evaluate FairQE under four evaluation settings related to gender bias. Experimental results demonstrate that FairQE produces more equitable evaluation behavior than existing QE baselines in both gender-ambiguous and gender-explicit conditions, while simultaneously achieving competitive or improved performance on general QE quality as measured against multidimensional quality metrics (MQM)-based benchmarks.

Our contributions are summarized as follows:
\begin{itemize}
    \item We propose FairQE, a multi-agent-based, fairness-aware QE framework that jointly addresses gender-ambiguous and gender-explicit scenarios.
    \item We introduce a bias-aware dynamic score aggregation mechanism that quantifies gender bias using gender-flipped translation variants and incorporates this information into QE scoring.
    \item Through extensive experiments across diverse gender bias evaluation settings and MQM-based QE benchmarks, we demonstrate that FairQE improves both fairness and overall evaluation performance compared to existing QE models.
\end{itemize}

\section{FairQE Framework}
\label{sec:method}
\begin{figure*}[t]
    \centering
    \includegraphics[width=0.9\textwidth]{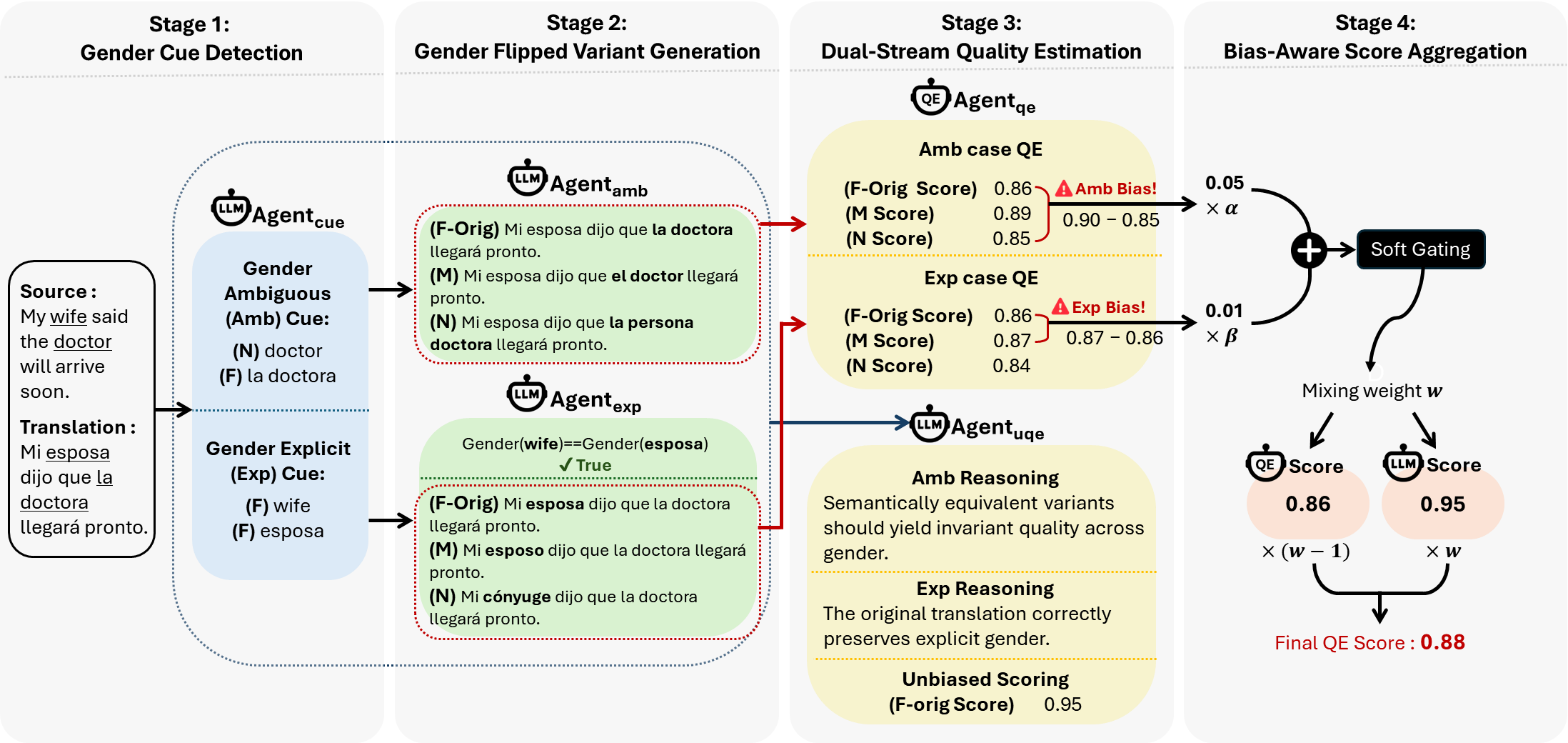}
    \caption{Overview of the proposed FairQE framework.
FairQE mitigates gender bias using four LLM-based agents for gender cue detection, variant generation, and bias-aware reasoning, in conjunction with a conventional QE module for quality scoring, yielding a fairer QE score. Here, F, M, and N represent Feminine, Masculine, and Neutral, respectively, and Orig refers to the original translation.}
    \label{fig:method}
\end{figure*}

FairQE is a multi-agent framework designed to mitigate gender bias in QE. As illustrated in Figure~\ref{fig:method}, the overall workflow is organized into four sequential stages: (1) Gender Cue Detection, (2) Gender-Flipped Variant Generation, (3) Dual-Stream Quality Estimation, and (4) Dynamic Bias-Aware Aggregation.

Across these stages, FairQE employs five agents: four LLM-based agents—the Gender Cue Detector ($Agent_{\textit{cue}}$), Gender-Ambiguous Variant Generator ($Agent_{\textit{amb}}$), Gender-Explicit Variant Generator ($Agent_{\textit{exp}}$), and Bias-Mitigating Quality Estimator ($Agent_{\textit{uqe}}$)—and a conventional QE model ($Agent_{\textit{qe}}$), which jointly combine LLM-based reasoning with traditional QE scoring. The details of each stage are described below.

\subsection{Gender Cue Detection}
The first stage serves as an initial verification module that identifies gender-related linguistic cues in a source--target sentence pair $(s, t)$. We employ a Gender Cue Detector ($Agent_{\textit{cue}}$) to determine whether the source sentence contains lexical cues, possibly spanning one or more words, that may induce gendered realizations in translation.

To ensure consistent and interpretable cue detection, we define an explicit taxonomy of gender bias cues to guide the behavior of $Agent_{\textit{cue}}$. Following prior work \citep{zaranis2025watching}, gender cues are first categorized into two primary types: Gender-Ambiguous and Gender-Explicit cues. Building on this distinction, we further decompose gender cues into twelve fine-grained categories that capture diverse linguistic sources of gender bias. These categories, denoted by $\mathcal{C}$, are summarized in Table~\ref{tab:tab7} and provided to $Agent_{\textit{cue}}$ through a prompt.

Based on this taxonomy, $Agent_{\textit{cue}}$ takes the cue taxonomy $\mathcal{C}$ as an explicit input and yields a set of detected gender cues, each linking a source cue span $c_s$ to its corresponding cue span in the target sentence $c_t$, along with its assigned cue category.

Formally, the output of this stage is defined as:
\begin{equation}
\resizebox{0.9\columnwidth}{!}{%
$\begin{split}
\mathcal{D} &= Agent_{\textit{cue}}(s, t, \mathcal{C}) \\
&= \{ (c_s^{(i)}, c_t^{(i)}) \mid
c_s^{(i)} \subset s,\;
c_t^{(i)} \subset t\}_{i=1}^{k}
\end{split}$%
}
\end{equation}

where $\mathcal{D}$ denotes the set of $k$ detected gender cue pairs and their assigned categories. If $\mathcal{D} = \emptyset$, the subsequent gender-flipped variant generation stages in Section~\ref{sec:method:gender-flipped-generation} are skipped to optimize computational efficiency.

\subsection{Gender-Flipped Variant Generation}
\label{sec:method:gender-flipped-generation}
At this stage, we generate gender-flipped variants of the original target sentence $t$ by applying modifications guided by the cues $\mathcal{D}$ detected by $Agent_{\textit{cue}}$.
The goal is to produce alternative gender realizations—Feminine~($F$), Masculine~($M$), and Neutral~($N$)—while preserving the original context and fluency. 
For instance, if $t$ contains a feminine cue, the agents generate valid masculine or neutral counterparts, ensuring they remain linguistically natural.

Given the distinct nature of gender-ambiguous and gender-explicit cues, we employ two specialized agents: the Gender-Ambiguous Variant Generator ($Agent_{\textit{amb}}$) and the Gender-Explicit Variant Generator ($Agent_{\textit{exp}}$). These agents operate selectively or in parallel depending on the cue category.

\paragraph{Gender-Ambiguous Variant Generator.} The $Agent_{\textit{amb}}$ handles cases where the gender of a cue in the source text is underspecified (corresponding to the ambiguous cue categories). 
Since the source gender is unknown, the goal of this agent is to generate all valid gender realizations that were not present in the original translation.

Formally, for the set of detected ambiguous cues $\mathcal{D}_{\textit{amb}}=\{(c_s,c_t)\in\mathcal{D}\mid \text{category}(c_s,c_t)\in\mathcal{C}_{\textit{amb}}\}$
, $Agent_{\textit{amb}}$ generates a set of gender-flipped variants $\mathcal{V}_{\textit{amb}}$:
\begin{equation}
\resizebox{0.9\columnwidth}{!}{$
\mathcal{V}_{\textit{amb}} =
\bigcup_{(c_s, c_t) \in \mathcal{D}_{\textit{amb}}}
Agent_{\textit{amb}}(s, t, c_s, c_t)
$}
\end{equation}
Each element in $\mathcal{V}_{\textit{amb}}$ is represented as a tuple $(s, t', g')$, where $t'$ denotes the target sentence modified to reflect a specific gender realization $g' \in \{F, M, N\}$.

\paragraph{Gender-Explicit Variant Generator.}
The $Agent_{\textit{exp}}$ handles cues where the source gender is explicitly marked.
Unlike the ambiguous case, the target gender is deterministically constrained to align
with the source.
Accordingly, the agent verifies whether the target realization $c_t$ satisfies the
source gender cue $c_s$, and generates gender-flipped variants for comparison.
The resulting alignment signal is used in the subsequent bias-mitigating reasoning stage.

Formally, for the set of detected explicit cues $\mathcal{D}_{\textit{exp}}=\{(c_s,c_t)\in\mathcal{D}\mid \text{category}(c_s,c_t)\in\mathcal{C}_{\textit{exp}}\}$, $Agent_{\textit{exp}}$ produces an alignment outcome
$\mathcal{A}^i \in \{\textit{True}, \textit{False}\}$ and a set of gender-flipped variants
$\mathcal{V}^i$ for each cue $(c_s, c_t)$:
\begin{equation}
\resizebox{0.6\columnwidth}{!}{$
(\mathcal{A}^i, \mathcal{V}^i) = Agent_{\textit{exp}}(s, t, c_s, c_t)
$}
\end{equation}

The final alignment signal $\mathcal{A}_{\textit{exp}}$ and the set of explicit variants
$\mathcal{V}_{\textit{exp}}$ are obtained by aggregating the cue-level outputs:
\begin{equation}
\resizebox{0.9\columnwidth}{!}{$
\mathcal{A}_{\textit{exp}} =
\bigwedge_{(c_s, c_t) \in \mathcal{D}_{\textit{exp}}} \mathcal{A}^i,
\qquad
\mathcal{V}_{\textit{exp}} =
\bigcup_{(c_s, c_t) \in \mathcal{D}_{\textit{exp}}} \mathcal{V}^i
$}
\end{equation}

\subsection{Dual-Stream Quality Estimation}
This stage aims to perform a multi-faceted assessment of translation quality. We employ a dual-stream approach that operates in parallel: quantitative scoring via a conventional QE model ($Agent_{\textit{qe}}$) and bias-mitigating reasoning via an LLM ($Agent_{\textit{uqe}}$). Both agents exploit the original sentence pair $(s, t)$ and the set of generated gender-flipped variants $\mathcal{V}$.

\paragraph{Quantitative Scoring via QE Model.}
The conventional QE model acts as a robust anchor for evaluating general fluency. $Agent_{\textit{qe}}$ computes quality scores for the original pair as well as for each generated gender-flipped variant. This allows us to measure not only the absolute quality but also the \textit{volatility} of the model's predictions under gender perturbations.

Formally, let $q_{\textit{orig}}$ be the score for $(s, t)$, and $\mathcal{Q}_{\textit{var}}$ be the set of scores for the set of variants $\mathcal{V} = \{(s'_i, t'_i)\}_{i=1}^N$:
\begin{equation}
\resizebox{0.75\columnwidth}{!}{$
\begin{split}
    q_{\textit{orig}} &= Agent_{\textit{qe}}(s, t), \\
    \mathcal{Q}_{\textit{var}} &= \{Agent_{\textit{qe}}(s', t') \mid (s', t') \in \mathcal{V} \}
\end{split}
$}
\end{equation}
If the model is biased, it is expected to assign significantly divergent scores to semantically equivalent variants despite the context remaining largely identical.

\paragraph{Bias-mitigating Reasoning via LLM.}
Unlike $Agent_{\textit{qe}}$, which treats inputs as a black box, $Agent_{\textit{uqe}}$ performs explicit reasoning to derive a debiased quality score $q_{\textit{uqe}}$.
The agent takes the detected cues $\mathcal{D}$, the set of variants $\mathcal{V}$, and the alignment signal $\mathcal{A}_{\textit{exp}}$ as context, applying distinct reasoning strategies based on the cue type:
\begin{itemize}
    \item \textbf{For Gender-Ambiguous Cues:} The agent checks for \textit{consistency}. It evaluates whether the translation quality remains invariant across valid gender realizations (F, M, N). Score disparities without contextual justification are flagged as bias.
    \item \textbf{For Gender-Explicit Cues:} The agent validates \textit{fidelity} by analyzing the generated variants alongside the deterministic alignment check $\mathcal{A}_{\textit{exp}}$. It contrasts the translation with the gender-flipped variants to confirm that the target gender strictly adheres to the source constraints, penalizing violations flagged by the verification signal.
\end{itemize}
Through this adaptive process, $Agent_{\textit{uqe}}$ outputs the final score:
\begin{equation}
\resizebox{0.7\columnwidth}{!}{$
q_{\textit{uqe}} = Agent_{\textit{uqe}}(s, t, \mathcal{D}, \mathcal{V}, \mathcal{A}_{\textit{exp}})
$}
\end{equation}

\subsection{Dynamic Bias-Aware Score Aggregation}
The final stage synthesizes the quantitative score from $Agent_\textit{qe}$ ($q_\textit{orig}$) and the reasoning-based score from $Agent_{\textit{uqe}}$ ($q_\textit{uqe}$) to derive the final quality score, $q_\textit{final}$.

Recent MQM-based studies suggest that while supervised QE models excel in fine-grained segment-level precision, LLMs demonstrate superior capabilities in reasoning-intensive tasks \cite{lu2024error}. Leveraging this complementarity, FairQE adopts a dynamic bias-aware score aggregation mechanism. Instead of static averaging, we use the QE model as a reliable anchor and dynamically increase the intervention of the LLM only when the severity of gender bias (bias score) warrants it.

\subsubsection{Quantifying Bias Scores}

We quantify bias score by analyzing the score distribution from $Agent_{qe}$ to compute the ambiguous bias and explicit bias.

\paragraph{Ambiguous Bias ($b_\textit{amb}$).} For gender-ambiguous cases, a bias-mitigating model should assign consistent scores across gender variations. We define volatility as the range between the maximum and minimum scores across the gender variants, and use it as the metric for bias.
\begin{equation}
\label{eq:b-amb}
\resizebox{0.9\columnwidth}{!}{$
b_{\textit{amb}}
= \max\!\bigl(\mathcal{Q}_\textit{var} \cup \{q_\textit{orig}\}\bigr)
 - \min\!\bigl(\mathcal{Q}_\textit{var} \cup \{q_\textit{orig}\}\bigr)
$}
\end{equation}

\paragraph{Explicit Bias ($b_\textit{exp}$).}
For gender-explicit cases, bias score is defined as a \emph{preference violation} when a
translation that violates the explicit gender constraint is scored higher than a
constraint-consistent alternative.
Let $\mathcal{A}_{\textit{exp}} \in \{\textit{True}, \textit{False}\}$ indicate whether the original
translation satisfies the explicit gender constraint.
If $\mathcal{A}_{\textit{exp}}=\textit{True}$, bias occurs when any constraint-violating variant
outperforms the original translation.
Otherwise, bias occurs when the original, constraint-violating translation is preferred
over any gender-flipped variant.

\begin{equation}
\resizebox{1.0\columnwidth}{!}{%
$
b_{\textit{exp}}
= \max
\begin{cases}
\max\!\Bigl(0,\; \max(\mathcal{Q}_\textit{var}) - q_{\textit{orig}} \Bigr)
& \parbox[t]{0.42\columnwidth}{if $\mathcal{A}_{\textit{exp}}=\textit{True}$} \\[6pt]
\max\!\Bigl(0,\; q_{\textit{orig}} - \max(\mathcal{Q}_\textit{var}) \Bigr)
& \parbox[t]{0.42\columnwidth}{if $\mathcal{A}_{\textit{exp}}=\textit{False}$}
\end{cases}
$%
}
\end{equation}

\subsubsection{Final Score Aggregation}
Here, $\alpha$ and $\beta$ control the relative contribution of $b_{\textit{amb}}$ and $b_{\textit{exp}}$, respectively. The total bias score $B$ is a weighted sum of the components, which determines the mixing weight $w$ via a soft-gating mechanism.
\begin{align}
\label{eq:B-def}
B &= \alpha \cdot b_{\textit{amb}} + \beta \cdot b_{\textit{exp}} \\[4pt]
\label{eq:w-def}
w &= \frac{B}{1 + B}, \quad (0 \le w < 1)
\end{align}

When bias is negligible ($B \approx 0$), $w$ approaches 0, prioritizing the segment-level precision of $Agent_\textit{qe}$. As bias score increases, $w$ grows, shifting reliance toward the bias-mitigating reasoning of $Agent_{\textit{uqe}}$. The final score is computed as:
\begin{equation}
\label{eq:final-score}
q_{\textit{final}}
= w \cdot q_\textit{uqe} + (1 - w) \cdot q_{\textit{orig}}
\end{equation}

\section{Experiments}
\label{sec:experiments}
\subsection{Experimental Setup}
\label{sec:exp-setup}

\begin{table*}[t]
\centering
\small
\renewcommand{\arraystretch}{1.25}
\setlength{\tabcolsep}{7.5pt}
\begin{tabular}{c|cccccc}
\Xhline{4\arrayrulewidth}
\textbf{Method} & \textbf{ES} & \textbf{FR} & \textbf{IT} & \textbf{AR} & \textbf{DE} & \textbf{HI} \\ \hline\hline
COMETKiwi 22 \cite{rei2022cometkiwi}
 & 0.9832 & \underline{0.9783} & 0.9791 & 0.9851 & 0.9937 & 0.9909 \\
COMETKiwi 23 XL \cite{rei2023scaling}
 & 0.9398 & 0.9028 & 0.9261 & 0.9841 & 0.9906 & 0.9840 \\
MetricX 24 L \cite{juraska2024metricx}
 & 0.9804 & 0.9714 & 0.9782 & 0.9623 & 0.9911 & 0.9945 \\
MetricX 24 XL \cite{juraska2024metricx}
 & 0.9802 & 0.9701 & 0.9816 & \textbf{0.9943} & \underline{0.9986} & \textbf{0.9989} \\
GEMBA-MQM \cite{kocmi2023gemba}
 & 0.9737 & 0.9658 & 0.9695 & 0.9700 & 0.9749 & 0.9740 \\
\rowcolor{gray!15}
FairQE (ours, w/ COMETKiwi 22)
 & \textbf{0.9947} & \textbf{0.9857} & \textbf{0.9917} & \underline{0.9938} & \textbf{0.9993} & \underline{0.9965} \\
\rowcolor{gray!15}
FairQE (ours, w/ MetricX 24 L)
 & \underline{0.9876} & 0.9731 & \underline{0.9881} & 0.9650 & 0.9954 & 0.9956 \\
\Xhline{4\arrayrulewidth}
\end{tabular}
\caption{Feminine-to-masculine QE score ratio on EN–* language pairs under gender-ambiguous (Fem. vs. Masc.) setting.
The best score for each language is shown in \textbf{bold}, and the second-best score is \underline{underlined}.}
\label{tab:tab1}
\end{table*}

\begin{table}[t]
\centering
\small
\resizebox{\columnwidth}{!}{
\begin{tabular}{c|ccc}
\Xhline{5\arrayrulewidth}
\textbf{Method} & \textbf{DE} & \textbf{ES} & \textbf{IT} \\ \hline\hline
COMETKiwi 22
 & 0.9737 & 0.9689 & 0.9694 \\
COMETKiwi 23 XL
 & 0.9643 & 0.9513 & 0.9436 \\
MetricX 24 L
 & \underline{0.9918} & \underline{0.9805} & 0.9737 \\
MetricX 24 XL
 & 0.9877 & 0.9756 & 0.9707 \\
GEMBA-MQM
 & 0.9820 & 0.9801 & \underline{0.9797} \\

\rowcolor{gray!15}
FairQE (ours, w/ COMETKiwi 22)
 & 0.9801 & 0.9693 & 0.9727 \\

\rowcolor{gray!15}
FairQE (ours, w/ MetricX 24 L)
 & \textbf{0.9921} & \textbf{0.9948} & \textbf{0.9884} \\
\Xhline{5\arrayrulewidth}
\end{tabular}
}
\caption{Neutral-to-gendered QE score ratio on EN–* language pairs under gender-ambiguous (Neutral vs. Gendered) setting. The best score for each language is shown in \textbf{bold}, and the second-best score is \underline{underlined}.}
\label{tab:tab2}
\end{table}

\begin{table}[t]
\centering
\small
\resizebox{\columnwidth}{!}{
\begin{tabular}{c|ccc}
\Xhline{5\arrayrulewidth}
\textbf{Method} & \textbf{AR} & \textbf{DE} & \textbf{HI} \\ \hline\hline
COMETKiwi 22
 & \underline{95.0} & \underline{99.2} & 55.3 \\
COMETKiwi 23 XL
 & \underline{95.0} & 98.7 & 73.1 \\
MetricX 24 L
 & 34.1 & 97.6 & 74.0 \\
MetricX 24 XL
 & 94.2 & 98.5 & \underline{78.9} \\
GEMBA-MQM
 & 88.5 & 94.0 & 72.0 \\
\rowcolor{gray!15}
FairQE (ours, w/ COMETKiwi 22)
 & \textbf{97.3} & \textbf{99.7} & 74.0 \\
\rowcolor{gray!15}
FairQE (ours, w/ MetricX 24 L)
 & 56.0 & 98.2 & \textbf{79.1} \\
\Xhline{5\arrayrulewidth}
\end{tabular}
}
\caption{Accuracy on EN–* language pairs under gender-explicit setting. The highest score for each language is shown in \textbf{bold}, and the second-highest score is \underline{underlined}.}
\label{tab:tab3}
\end{table}

We evaluate FairQE under two complementary experimental settings:
(1) \textbf{Gender fairness evaluation}, which assesses whether existing gender bias is effectively mitigated,
and (2) \textbf{QE performance evaluation}, which assesses whether FairQE maintains general-purpose QE performance.
All datasets follow an EN--* language-pair setting.

\paragraph{Gender Fairness Evaluation.}
Gender bias is evaluated using setting-specific criteria tailored to the nature of gender ambiguity and explicitness.
Dataset details for this evaluation and formal definitions of all evaluation metrics are provided in Appendix~\ref{app:datasets} and Appendix~\ref{app:eval_metric_app}, respectively.

\vspace{4pt}
\begin{table*}[t]
\centering
\scriptsize
\renewcommand{\arraystretch}{0.9} 
\resizebox{0.80\textwidth}{!}{ 
\begin{tabular}{
>{\centering\arraybackslash}m{3.8cm} |
>{\centering\arraybackslash}m{1.6cm} |
cc | cc
}
\Xhline{5\arrayrulewidth}
\multirow{2}{*}{\textbf{Method}}
& \multirow{2}{*}{\textbf{avg-corr}}
& \multicolumn{2}{c|}{\textbf{System-Level}}
& \multicolumn{2}{c}{\textbf{Segment-Level}} \\
&  & accuracy & pearson & acc-t & pearson \\
\hline\hline
\noalign{\vskip 1pt}
BLEU & 0.742 & 0.894 & 0.917 & 0.520 & 0.192 \\
chrF & 0.722 & 0.818 & 0.866 & 0.519 & 0.232 \\
BERTScore & 0.754 & 0.879 & 0.891 & 0.528 & 0.325 \\
COMETKiwi 22 & 0.743 & 0.864 & 0.901 & \underline{0.548} & 0.224 \\
COMETKiwi 23 XL & 0.764 & 0.909 & 0.900 & 0.541 & 0.308 \\
MetricX 24 L & 0.734 & 0.894 & 0.940 & 0.493 & 0.155 \\
MetricX 24 XL & 0.750 & 0.848 & 0.872 & 0.528 & 0.378 \\
GEMBA-MQM & \textbf{0.830} & \underline{0.970} & \textbf{0.981} & \textbf{0.574} & \textbf{0.568} \\
\rowcolor{gray!15}
FairQE (ours, w/ COMETKiwi 22)
& \underline{0.812} & \textbf{0.985} & 0.950 & \textbf{0.574} & 0.424 \\
\rowcolor{gray!15}
FairQE (ours, w/ MetricX 24 L)
& 0.806 & \underline{0.970} & \underline{0.953} & 0.545 & \underline{0.463} \\
\Xhline{5\arrayrulewidth}
\end{tabular}
}
\caption{Results on the WMT 2023 Metrics Shared Task for EN--DE at system-level and segment-level evaluation. The highest score in each column is shown in \textbf{bold}, and the second-highest score is \underline{underlined}.}
\label{tab:tab4}
\end{table*}
\vspace{4pt}

\begin{itemize}
    \item \textbf{Gender-ambiguous (Fem. vs. Masc.):}
    We compare semantically equivalent feminine and masculine translations for gender-ambiguous sources.
    Fairness is evaluated using the feminine-to-masculine QE score ratio, where values closer to 1 indicate less gender preference. We conduct this evaluation using the GATE \cite{rarrick2023gate} dataset and the contextual subset of MT-GenEval \cite{currey2022mtgeneval}, with preceding context removed to preserve gender ambiguity.
    \item \textbf{Gender-ambiguous (Neutral vs. Gendered):}
    We evaluate whether a QE model prefers gender-neutral translations over gender-specific ones for gender-ambiguous sources. Fairness is assessed using the neutral-to-gendered QE score ratio, where values greater than 1 indicate a preference for preserving gender ambiguity. We conduct this evaluation using the mGeNTE \cite{savoldi2025mgente} dataset.
    \item \textbf{Gender-explicit:}
    We assess whether the QE model assigns higher scores to gender-aligned translations than to gender-misaligned ones for gender-explicit sources.
    Performance is measured using binary accuracy, where a prediction is considered correct if the gender-aligned translation receives a higher QE score. We conduct this evaluation using the counterfactual subset of MT-GenEval.
\end{itemize}

\paragraph{QE Performance Evaluation.}
We assess QE performance following the official WMT 2023 Metrics Shared Task \cite{freitag2023wmtmetrics} setup.
We evaluate on the EN--DE language pair using MQM ratings, covering 14 MT systems with 557 segments per system.

\paragraph{Baselines.}
We compare FairQE against regression-based QE models, including COMETKiwi~22 \cite{rei2022cometkiwi}, COMETKiwi~23 XL \cite{rei2023scaling}, MetricX~24~L \cite{juraska2024metricx}  and MetricX~24~XL \cite{juraska2024metricx}. 
We also include GEMBA-MQM \cite{kocmi2023gemba}, an LLM-based evaluator that performs MQM-style error analysis,
instantiated with \texttt{gpt-4.1-mini}\footnote{\url{https://platform.openai.com/docs/models/gpt-4.1-mini}}.
For QE performance evaluation, we additionally include BERTScore \cite{zhang2020bertscore}, BLEU \cite{papineni2002bleu}, and ChrF \cite{popovic2015chrf} as baselines,
enabling comparisons with both learned QE models and traditional automatic evaluation metrics.

\paragraph{Models and Implementation.}
For all experiments, including both gender bias mitigation and QE performance evaluation,
we use \texttt{gpt-4.1-mini} for all LLM-based agents.
As the underlying QE backbone, we employ COMETKiwi~22 and MetricX~24~L. 
For the aggregation hyperparameters in Equation~(\ref{eq:B-def}), we set $\alpha=\beta=5$ in all experiments unless otherwise stated.

\begin{figure*}[t]
    \centering
    \includegraphics[width=0.97\textwidth]{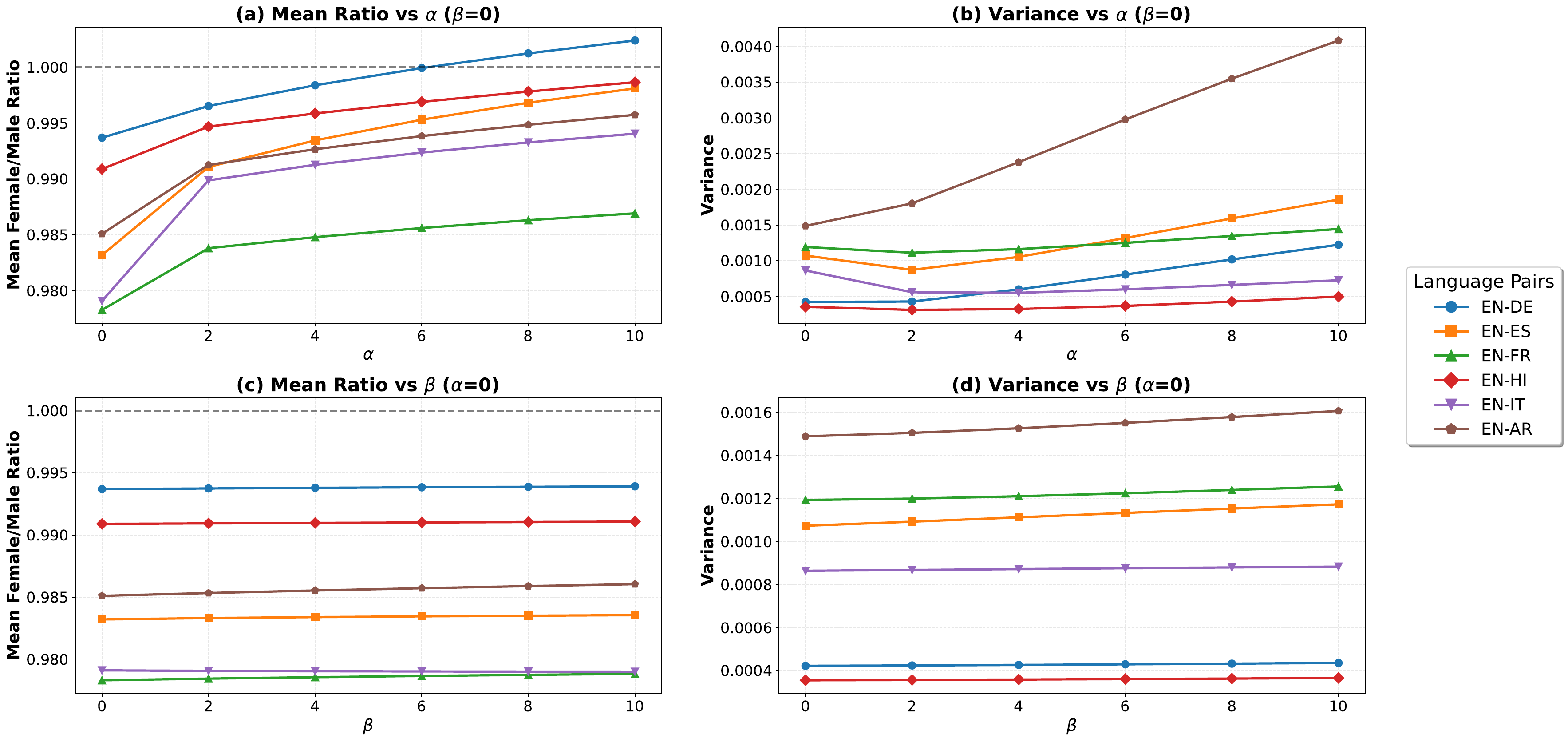}
    \caption{Analysis of hyperparameters $\alpha$ and $\beta$ across six language pairs under gender-ambiguous (Fem. vs. Masc.) setting.
Panels (a) and (c) show the mean feminine-to-masculine QE score ratio, while panels (b) and (d) report the variance of QE scores as the hyperparameter value increases.}
    \label{fig:alpha_beta_effect_amb_fm}
\end{figure*}

\subsection{Experimental Results}
\label{sec:exp_results}

\subsubsection{Gender-ambiguous (Fem. vs. Masc.)}
\label{sec:amb_fm_results}
Table~\ref{tab:tab1} reports the results for this setting.
Across most language pairs, all baseline QE models exhibit a feminine-to-masculine QE score ratio below 1, indicating a systematic masculine bias that favors masculine translations even when the source sentence contains no gender cues.

Under the same conditions, FairQE (w/ COMETKiwi~22) achieves the closest performance to parity in four out of six language pairs (EN--ES, EN--FR, EN--IT, EN--DE), and attains the second-best performance on the remaining two pairs (EN--AR, EN--HI).
In addition, FairQE with MetricX~24~L as the QE backbone consistently outperforms its corresponding backbone across all language pairs.
These results demonstrate that FairQE consistently restores gender balance across diverse language settings.

\subsubsection{Gender-ambiguous (Neutral vs. Gendered)}
\label{sec:amb_ng_results}
Table~\ref{tab:tab2} reports the results for this setting.
Baseline QE models generally fail to sufficiently favor gender-neutral translations, indicating a bias toward unnecessary gender assignment even in the absence of explicit gender cues.

In contrast, FairQE (w/ MetricX~24~L) achieves the best performance across all language pairs (EN--DE, EN--ES, EN--IT), while FairQE (w/ COMETKiwi~22) consistently improves over its backbone model.
These results indicate that FairQE functions as a model-agnostic framework for improving gender-neutral evaluation.

\subsubsection{Gender-explicit}
\label{sec:exp_results}
Table~\ref{tab:tab3} reports the results for this setting. FairQE (w/ COMETKiwi~22) achieves the highest accuracy on two out of three language pairs (EN--AR, EN--DE), and attains the second-best performance on the remaining pair (EN--HI). This corresponds to an improvement of 18.7 percentage points over its underlying QE backbone, COMETKiwi~22, demonstrating that FairQE effectively corrects erroneous QE judgments even when gender cues are explicitly specified.

Meanwhile, GEMBA-MQM, a representative LLM-as-judge–based QE method, achieves an average accuracy of approximately 84\% across the three language pairs, which is notably lower than the approximately 90\% accuracy achieved by FairQE (w/ COMETKiwi~22). This observation is consistent with prior findings that gender bias inherent to LLMs can propagate into the evaluation stage \cite{chen2024humans}. In contrast, by performing explicit reasoning over multiple gender-flipped variants, FairQE provides more reliable gender judgments in gender-explicit scenarios.

\subsubsection{QE Performance}
\label{sec:qe_results}
Experimental results in Table~\ref{tab:tab4} show that the proposed method achieves the best performance in terms of accuracy and accuracy-t at both the system and segment levels. In contrast, it obtains a slightly lower score on the overall avg-corr metric (0.812) compared to GEMBA-MQM (0.830). However, considering that GEMBA-MQM exhibited relatively substantial gender bias in the gender bias experiments, this result can be interpreted as a meaningful trade-off between fairness and evaluation performance.
Moreover, FairQE with COMETKiwi~22 as the QE backbone substantially outperforms its backbone model, improving the avg-corr from 0.743 to 0.812. This demonstrates that our approach not only mitigates gender bias in existing QE models but also improves overall QE performance by effectively integrating more neutral LLM-based evaluation signals.

\subsection{Ablation Studies}
\label{sec:ablations}
\subsubsection{Effect of $\alpha$ and $\beta$ on Fairness and Stability}
\label{sec:ablation_hyper_params}

\begin{table*}[t]
\centering
\small
\renewcommand{\arraystretch}{1.25}
\setlength{\tabcolsep}{7.5pt}
\begin{tabular}{c|ccccccc}
\Xhline{4\arrayrulewidth}
\textbf{Method} & \textbf{ES} & \textbf{FR} & \textbf{IT} & \textbf{AR} & \textbf{DE} & \textbf{HI} & \textbf{Avg.} \\ \hline\hline
COMETKiwi 22
 & 0.9832 & 0.9783 & 0.9791 & 0.9851 & \underline{0.9937} & 0.9909 & 0.9851 \\
GEMBA-MQM
 & 0.9737 & 0.9658 & 0.9695 & 0.9700 & 0.9749 & 0.9740 & 0.9713 \\
\rowcolor{gray!15}
FairQE-NoVar (Ours)
 & \underline{0.9892} & \underline{0.9826} & \underline{0.9899} & 0.9925 & 0.9928 & \underline{0.9921} & \underline{0.9899} \\
\rowcolor{gray!15}
FairQE-UQEOnly (Ours)
 & 0.9812 & 0.9799 & 0.9777 & \textbf{0.9951} & 0.9879 & 0.9774 & 0.9832 \\
\rowcolor{gray!15}
FairQE (Ours)
 & \textbf{0.9947} & \textbf{0.9857} & \textbf{0.9917} & \underline{0.9938} & \textbf{0.9993} & \textbf{0.9965} & \textbf{0.9936} \\
\Xhline{4\arrayrulewidth}
\end{tabular}
\caption{Component-wise ablation analysis of the FairQE framework on EN--* language pairs under the gender-ambiguous (Fem. vs. Masc.) setting. FairQE-NoVar removes gender-flipped variant generation (Stage 2), while FairQE-UQEOnly uses only the UQE-based score ($q_{uqe}$).}
\label{tab:ablation_results}
\end{table*}

We analyze the impact of the hyperparameters $\alpha$ and $\beta$. Figure~\ref{fig:alpha_beta_effect_amb_fm} presents the feminine-to-masculine QE score ratio and variance under the
\textbf{Gender-ambiguous (Fem. vs. Masc.)} setting as $\alpha$ and $\beta$ vary. We evaluate both hyperparameters from 0 to 10 in increments of 2.

As shown in panel (a), the ratio gradually increases across all language pairs as $\alpha$ increases.
This indicates that the previously masculine-biased score distribution moves closer to the ideal value of 1,
or may shift toward feminine bias when $\alpha$ becomes sufficiently large.
This behavior is not due to differences in score ranges, but rather to the tendency of LLM-based score inference
to produce relatively larger score magnitudes (or score differences) than the base QE model,
even within the same scoring scale.

Regarding variance, panel (b) shows that most language pairs except EN--AR exhibit a slight decrease at smaller $\alpha$ values
followed by an increase as $\alpha$ grows.
This shows that even when the ratio approaches 1 and gender bias is mitigated,
setting $\alpha$ excessively large can increase score instability,
highlighting the importance of proper hyperparameter selection for $\alpha$.

The analysis of the hyperparameter $\beta$ is shown in panels (c) and (d). Unlike $\alpha$, changes in $\beta$ result in largely similar levels of feminine-to-masculine QE score ratio and variance.
This is because the experiment uses gender-ambiguous source inputs,
for which the influence of $\beta$, which operates on explicit gender cues, is limited.
In contrast, under the gender-explicit setting, the effect of $\beta$ becomes more pronounced,
while the influence of $\alpha$ is relatively reduced (See Appendix~\ref{app:hyperparameter} for additional details.).

Overall, these ablation results show that $\alpha$ plays a more important role in gender-ambiguous settings,
while $\beta$ is more influential in gender-explicit settings, confirming that FairQE operates in accordance with its design intent.

\subsubsection{Component-wise Ablation Analysis} 
Table~\ref{tab:ablation_results} presents the component-wise ablation results of FairQE under the gender-ambiguous (Fem. vs. Masc.) setting. In this experiment, FairQE uses COMETKiwi 22 and gpt-4.1-mini as backbones. We analyze how the performance improvements stem from \textbf{(1) the gender-flipped variant generation (Stage 2)} and \textbf{(2) the LLM-based evaluation component and its aggregation with traditional QE metrics}.

\paragraph{Effect of Gender-Flipped Variant Generation (FairQE-NoVar).} 
To assess the contribution of Stage 2, we consider FairQE-NoVar, which removes gender-flipped variant generation. In this setting, the detected gender cues, source, and translation are directly provided to the LLM without contrastive comparison. FairQE-NoVar outperforms both COMETKiwi 22 and GEMBA-MQM in most language pairs (except EN-DE), but consistently underperforms the full FairQE model. This indicates that cue-guided LLM evaluation alone provides improvements, while contrastive comparison via gender-flipped variants yields additional gains.

\paragraph{Effect of the LLM-based Evaluator ($q_{uqe}$)} 
To examine the effectiveness of the LLM-based evaluation component, we consider the FairQE-UQEOnly setting, where only the LLM-based score ($q_{uqe}$) is used. The results show that the standalone $q_{uqe}$ achieves competitive performance, outperforming COMETKiwi 22 for some language pairs, while underperforming it for others. These results suggest that the LLM-based evaluator does not consistently provide superior performance across all settings. 

In contrast, the full FairQE model, which incorporates Bias-Aware Score Aggregation, achieves the best performance across all language pairs except EN–AR. This demonstrates that dynamically combining signals from the LLM-based metric and the traditional QE metric yields more stable and reliable performance than relying on either component alone.

\section{Related Works}
\label{sec:relatedworks}
QE seeks to assess the quality of MT outputs without relying on reference translations, and has emerged as a practical complement to reference-based MT evaluation \cite{zhao2024handcrafted}. A dominant line of research in QE has focused on regression-based models that leverage pretrained language models as encoders. These approaches typically combine contextualized representations with task-specific prediction heads to estimate translation quality, and have demonstrated strong correlations with human judgments across diverse language pairs and domains \cite{rei2020comet, rei2022cometkiwi, juraska2024metricx}.

Beyond conventional regression-based QE models, recent work has explored LLM-as-a-judge approaches for QE. Rather than relying on explicit regression objectives, these methods leverage the reasoning and instruction-following capabilities of LLMs to evaluate translation quality through prompt-based inference, often from multiple complementary perspectives. Several studies incorporate MQM \cite{lommel2013multidimensional, freitag2021experts} guidelines or fine-grained error taxonomies directly into prompts to structure and guide the evaluation process \cite{kocmi2023gemba, lu2024error}. In addition, recent work has introduced multi-agent debate mechanisms to further enhance QE robustness and coverage \cite{feng2025m}. Together, these advances reflect a growing interest in interpretable, reasoning-driven evaluation paradigms for QE.

Meanwhile, prior research has shown that QE models can encode various linguistic and social biases, with gender bias emerging as a particularly salient concern. Existing studies report that QE models may systematically favor specific gender realizations in gender-ambiguous source sentences, or assign higher scores to gender-mismatched translations even when the source sentence is gender-explicit \cite{savoldi2021gender}. To investigate these phenomena, the MT community has conducted extensive analyses of score distributions across gender conditions, compared bias patterns across different QE architectures, and proposed benchmarks and evaluation protocols to quantify gender-related biases in a controlled manner \cite{zaranis2025watching, filandrianos2025gambit, mastromichalakis2025assumed}.

While these studies provide valuable insights into diagnosing and comparing gender bias in QE models, methodological approaches that directly mitigate gender bias at the QE evaluation stage remain underexplored. This work addresses this gap by proposing a new direction for systematically considering and improving gender fairness at the QE evaluation stage.

\section{Conclusion}
\label{sec:conclusion}
We propose FairQE, a multi-agent, fairness-aware framework for quality estimation that mitigates gender bias in both gender-ambiguous and gender-explicit scenarios. FairQE detects gender cues, generates gender-flipped variants, and dynamically combines conventional QE scores with LLM-based unbiased reasoning using a bias-aware aggregation mechanism. Experimental results across diverse gender bias evaluation settings show that FairQE consistently improves gender fairness over strong QE baselines, while maintaining or improving general QE performance under MQM-based meta-evaluation. These findings demonstrate that gender bias in QE can be mitigated without sacrificing evaluation accuracy, making FairQE a practical and model-agnostic solution for fairer MT evaluation.

\section*{Limitations}
FairQE relies on LLM-based components for unbiased quality estimation, which may introduce variability depending on the underlying LLM and decoding configuration. To mitigate this issue, all LLM-based components are executed with deterministic decoding and fixed prompts, and the final quality score is anchored to the underlying QE backbone, with the contribution of the LLM-based unbiased estimator increased only when the estimated bias score is high. 

Errors in gender cue detection may propagate to downstream stages; however, the benchmarks used in this study are constructed to explicitly contain gender-related phenomena, and FairQE consistently reduces bias relative to the corresponding QE backbones across all experimental settings, indicating stable behavior of the overall pipeline in practice. 

The bias aggregation hyperparameters may be sensitive to the choice of QE backbone or dataset, but we use a single fixed setting across all experiments and observe consistent improvements, suggesting reasonable stability and generality. 

Finally, our experiments rely on API-based LLMs to ensure reliable instruction-following behavior and do not include evaluations with open-source LLMs; nevertheless, FairQE is designed to be model-agnostic with respect to the choice of LLM, and extending the evaluation to open-source models is left for future work.

\section*{Ethics Statement}
This work aims to improve fairness in machine translation quality estimation by mitigating gender-related biases in existing evaluation models, particularly in gender-ambiguous and gender-explicit contexts. All experiments are conducted on publicly available benchmarks, and we do not collect or infer sensitive personal attributes beyond the explicit linguistic gender phenomena encoded in the data.

FairQE relies on LLMs for bias-aware reasoning. While LLMs may themselves encode social biases, our framework is explicitly designed to detect and mitigate such biases at the evaluation stage rather than amplify them, without enforcing a single normative notion of gender usage. Finally, FairQE is intended as an offline evaluation and analysis framework, and should be applied with appropriate human oversight in sensitive or high-stakes settings.

\section*{Acknowledgements}
This work was supported by the Institute of Information \& Communications Technology Planning \& Evaluation (IITP) grant funded by the Korea government (MSIT) [RS-2021-II211341, Artificial Intelligence Graduate School Program (Chung-Ang University)] and by the National Research Foundation of Korea (NRF) grant funded by the Korea government (MSIT) (RS-2025-00556246).

\bibliography{custom}

\clearpage
\appendix

\section{Experimental Details}
\subsection{Models}

To improve readability, the following abbreviations are used in our experiment tables to denote specific evaluation models:
COMETKiwi 22 (Unbabel/wmt22-cometkiwi-da), 
COMETKiwi 23 XL (Unbabel/wmt23-cometkiwi-da-xl), 
MetricX 24 L (google/metricx-24-hybrid-large-v2p6), 
MetricX 24 XL (google/metricx-24-hybrid-xl-v2p6), and 
GEMBA-MQM (instantiated with GPT-4.1-mini).

\subsection{Datasets}
\label{app:datasets}
This study evaluates the effectiveness of FairQE with respect to gender ambiguity and gender explicitness in source sentences, following the experimental protocols of prior work and using three benchmark datasets.

MT-GenEval \cite{currey2022mtgeneval} is a Wikipedia-based benchmark for gender bias evaluation. We use two subsets: the Contextual Subset, in which the preceding context is removed to preserve gender ambiguity in the source sentences, and the Counterfactual Subset, which evaluates explicit gender distinctions using contrastive pairs such as “He/She is a doctor.” Among the eight available target languages (AR, DE, ES, FR, HI, IT, PT, RU), we conduct experiments on AR, DE, and HI.

GATE \cite{rarrick2023gate} is a linguistically designed corpus that contains single gender-marked entities, and we evaluate on the ES, FR, and IT language pairs.

mGeNTE \cite{savoldi2025mgente} includes gender-neutral translations produced by professional translators. We use the gender-ambiguous source set (Set-N) to analyze whether models prefer neutral expressions over unnecessary gender assignments. The target languages are ES, DE, and IT.

All datasets are provided in EN–* language-pair settings. These datasets are freely available for research use under the MIT, CC-BY-SA-3.0, and CC-BY-4.0 licenses, respectively.

\subsection{Evaluation Metrics on Gender Fairness}
\label{app:eval_metric_app}

\subsubsection{Gender-Ambiguous: Feminine vs. Masculine}

For a gender-ambiguous source sentence $s$, we compare a feminine translation $h_F$ and a masculine translation $h_M$, which are semantically equivalent except for gender realization.
We compute the relative score ratio:
\begin{equation}
    r_{\text{m/f}}(s) = \frac{\text{QE}(s, h_F)}{\text{QE}(s, h_M)}.
\end{equation}
An ideal fair QE model yields $r_{\text{m/f}}(s) = 1$, indicating no preference for either gender form.

\subsubsection{Gender-Ambiguous: Neutral vs. Gendered}

For gender-ambiguous sources, we additionally compare a gender-neutral translation $h_N$ with a gender-specific translation $h_G$.
The relative preference for neutrality is measured as:
\begin{equation}
    r_{\text{neutral}}(s) = \frac{\text{QE}(s, h_N)}{\text{QE}(s, h_G)}.
\end{equation}
Values greater than 1 indicate a preference for preserving gender ambiguity through neutral translations.

\subsubsection{Gender-Explicit Accuracy}

For gender-explicit source sentences, we compare a gender-aligned translation $h^{\text{corr}}$ with a gender-misaligned translation $h^{\text{incorr}}$.
Following prior work, we evaluate performance using binary accuracy:
\begin{equation}
\begin{aligned}
\text{Acc}_{\text{explicit}}(S) = \\\frac{1}{|S|} \sum_{s_G \in S} \mathbb{I} \Big[ & \text{QE}(s_G, h^{\text{corr}})
 > \text{QE}(s_G, h^{\text{incorr}}) \Big].
\end{aligned}
\end{equation}

where $\mathbb{I}[\cdot]$ is the indicator function.

\subsection{Hardware Specification}
All our experiments were conducted using a single NVIDIA A100 GPU.

\label{sec:A_app}

\section{Additional Results}
\begin{table*}[t]
\centering
\normalsize
\setlength{\tabcolsep}{2.5pt} 
\resizebox{\textwidth}{!}{
\begin{tabular}{l | cc | cc | cc | cc | cc | cc} 
\Xhline{5\arrayrulewidth}
\multicolumn{1}{c|}{\textbf{Method}}
& \multicolumn{2}{c|}{\textbf{ES}} 
& \multicolumn{2}{c|}{\textbf{FR}}
& \multicolumn{2}{c|}{\textbf{IT}}
& \multicolumn{2}{c|}{\textbf{AR}}
& \multicolumn{2}{c|}{\textbf{DE}}
& \multicolumn{2}{c}{\textbf{HI}} \\
& Avg. & Std. & Avg. & Std. & Avg. & Std. & Avg. & Std. & Avg. & Std. & Avg. & Std. \\
\hline\hline
CometKiwi 22
& 0.9832 & \underline{0.0332}
& \underline{0.9783} & \textbf{0.0348}
& 0.9791 & 0.0543
& 0.9851 & \textbf{0.0470}
& 0.9937 & \underline{0.0240}
& 0.9909 & 0.0234 \\

CometKiwi 23 XL
& 0.9398 & 0.1089
& 0.9028 & 0.1484
& 0.9261 & 0.1037
& 0.9841 & 0.0910
& 0.9906 & 0.0544
& 0.9840 & 0.0498 \\

MetricX 24 L
& 0.9804 & 0.0381
& 0.9714 & 0.0545
& 0.9782 & 0.0433
& 0.9623 & 0.0580
& 0.9911 & 0.0285
& 0.9945 & 0.0275 \\

MetricX 24 XL
& 0.9802 & 0.0454
& 0.9701 & 0.0589
& \underline{0.9816} & 0.0517
& \textbf{0.9943} & 0.0517
& \underline{0.9986} & \textbf{0.0241}
& \textbf{0.9989} & \textbf{0.0172} \\

GEMBA-MQM
& 0.9737 & 0.0460
& 0.9658 & 0.0404
& 0.9695 & \underline{0.0285}
& 0.9700 & 0.0641
& 0.9749 & 0.0508
& 0.9740 & 0.0407 \\

\rowcolor{gray!15}
FairQE (CK22)
& \textbf{0.9947} & 0.0347
& \textbf{0.9857} & \underline{0.0356}
& \textbf{0.9917} & \textbf{0.0244}
& \underline{0.9938} & \underline{0.0523}
& \textbf{0.9993} & 0.0265
& \underline{0.9965} & \underline{0.0186} \\

\rowcolor{gray!15}
FairQE (MX24L)
& \underline{0.9876} & \textbf{0.0304}
& 0.9731 & 0.0517
& 0.9881 & 0.0441
& 0.9650 & 0.0581
& 0.9954 & 0.0327
& 0.9956 & 0.0342 \\
\Xhline{5\arrayrulewidth}
\end{tabular}
}
\caption{Average (Avg.) and standard deviation (Std.; lower is better) of the feminine-to-masculine QE score ratio on EN–* language pairs under gender-ambiguous (Fem. vs. Masc.) setting.
The best score for each language is shown in \textbf{bold}, and the second-best score is \underline{underlined}.}
\label{tab:tab5}
\end{table*}
\begin{table*}[t]
\centering
\small
\renewcommand{\arraystretch}{1.25}
\setlength{\tabcolsep}{7.5pt}
\begin{tabular}{c|ccccccc}
\Xhline{4\arrayrulewidth}
\textbf{Method} & \textbf{ES} & \textbf{FR} & \textbf{IT} & \textbf{AR} & \textbf{DE} & \textbf{HI} & \textbf{Avg.} \\ \hline\hline
COMETKiwi 22
 & 0.9832 & 0.9783 & 0.9791 & 0.9851 & 0.9937 & 0.9909 & 0.9851 \\
\rowcolor{gray!15}
FairQE (ours, w/ COMETKiwi 22)
 & \textbf{0.9947} & \textbf{0.9857} & \textbf{0.9917} & \textbf{0.9938} & \textbf{0.9993} & \textbf{0.9965} & \textbf{0.9936} \\
COMETKiwi 23 XL
 & 0.9398 & 0.9028 & 0.9261 & 0.9841 & 0.9906 & 0.9840 & 0.9546 \\
\rowcolor{gray!15}
FairQE (ours, w/ COMETKiwi 23 XL)
 & \textbf{0.9718} & \textbf{0.9271} & \textbf{0.9564} & \textbf{0.9924} & \textbf{1.0224} & \textbf{1.0100} & \textbf{0.9800} \\
\Xhline{4\arrayrulewidth}
\end{tabular}
\caption{Feminine-to-masculine QE score ratio on EN--* language pairs under the gender-ambiguous (Fem. vs. Masc.) setting. For each backbone, the better score between the baseline and FairQE is highlighted in \textbf{bold}.}
\label{tab:backbone_results}
\end{table*}
\begin{table*}[t]
\centering
\small
\renewcommand{\arraystretch}{1.25}
\setlength{\tabcolsep}{7.5pt}
\begin{tabular}{c|cccccc}
\Xhline{4\arrayrulewidth}
\textbf{Method} & \textbf{ES} & \textbf{FR} & \textbf{IT} & \textbf{AR} & \textbf{DE} & \textbf{HI} \\ \hline\hline
GEMBA-MQM (w/ gpt-4.1-mini)
 & 0.9737 & 0.9658 & 0.9695 & 0.9700 & 0.9749 & 0.9740 \\
GEMBA-MQM (w/ o3)
 & 0.9738 & 0.9723 & 0.9660 & 0.9767 & 0.9772 & 0.9762 \\
\rowcolor{gray!15}
FairQE (ours)
 & \textbf{0.9947} & \textbf{0.9857} & \textbf{0.9917} & \textbf{0.9938} & \textbf{0.9993} & \textbf{0.9965} \\
\Xhline{4\arrayrulewidth}
\end{tabular}
\caption{Feminine-to-masculine QE score ratio on EN--* language pairs in the gender-ambiguous (Fem. vs. Masc.) setting. FairQE (ours) uses COMETKiwi 22 and \texttt{gpt-4.1-mini} as backbones.}
\label{tab:strong_llm}
\end{table*}

\begin{table}[t]
\centering
\small
\resizebox{\columnwidth}{!}{
\begin{tabular}{c|ccc}
\Xhline{3\arrayrulewidth}
\textbf{Detected Cue Type} & \textbf{Count} & \textbf{Proportion (\%)} & \textbf{Ratio} \\ \hline\hline
Gender Ambiguous & 1,007 & 67.0 & 1.0033 \\
Gender Explicit & 324 & 21.6 & 0.9925 \\
Both & 9 & 0.6 & 0.9836 \\
None & 163 & 10.8 & 0.9889 \\
\Xhline{3\arrayrulewidth}
\end{tabular}
}
\caption{Cue detection distribution and feminine-to-masculine QE score ratio analysis in the gender-ambiguous setting (EN--DE).}
\label{tab:cue_ambiguous}
\end{table}

\begin{table}[t]
\centering
\small
\resizebox{\columnwidth}{!}{
\begin{tabular}{c|ccc}
\Xhline{3\arrayrulewidth}
\textbf{Detected Cue Type} & \textbf{Count} & \textbf{Proportion (\%)} & \textbf{Accuracy} \\ \hline\hline
Gender Explicit & 1,093 & 91.1 & 99.9 \\
Gender Ambiguous & 27 & 2.3 & 100 \\
Both & 28 & 2.3 & 96.4 \\
None & 52 & 4.3 & 96.2 \\
\Xhline{3\arrayrulewidth}
\end{tabular}
}
\caption{Cue detection distribution and accuracy in the gender-explicit setting (EN--DE).}
\label{tab:cue_explicit}
\end{table}
\begin{table*}[t]
\centering
\small
\renewcommand{\arraystretch}{1.2}
\setlength{\tabcolsep}{5pt}
\resizebox{\textwidth}{!}{
\begin{tabular}{c|p{6.5cm}|c|c|c}
\Xhline{3\arrayrulewidth}
\textbf{Type} & \textbf{Sentence} & \textbf{$q_{uqe}$} & \textbf{$q_{orig}$} & \textbf{FairQE Score} \\ \hline\hline
Source 
& It has been two days and I have to think about telling all those adventurers. 
& -- & -- & -- \\ \hline
Target (Masc.) 
& Lleva dos días y tengo que ir pensando en decírselo a todos esos aventureros. 
& 95 & 0.7605 & 0.7605 \\ \hline
Target (Fem.) 
& Lleva dos días y tengo que ir pensando en decírselo a todas esas aventureras. 
& 95 & 0.7467 & 0.8420 \\
\Xhline{3\arrayrulewidth}
\end{tabular}
}
\caption{A failure case in the gender-ambiguous (Fem. vs. Masc.) setting for EN--ES.}
\label{tab:failure_case}
\end{table*}

\subsection{Experimental Results with Variance}
Table~\ref{tab:tab5} reports the results under the gender-ambiguous (Fem. vs. Masc.) source setting.
Overall, the variance does not increase substantially compared to the baseline, and for certain language pairs (EN–ES and EN–IT), it even decreases, while the average feminine-to-masculine ratio remains close to 1, indicating reduced gender bias.
Although a slight increase in variance can occur due to differences in the score scale between the agent-based QE outputs and the underlying QE model predictions, this does not negatively impact overall performance.
In fact, despite marginally higher variance, our method consistently outperforms lower-variance baselines across all three gender bias evaluation settings (see Section \ref{sec:amb_fm_results}--\ref{sec:exp_results}) as well as in general QE performance evaluation (see Section \ref{sec:qe_results}), suggesting that the observed variance increase is acceptable in practice.

\subsection{Hyperparameter Analysis on Gender-explicit Settings}
\label{app:hyperparameter}
\begin{figure}[!ht]
    \centering
    \includegraphics[width=\columnwidth]{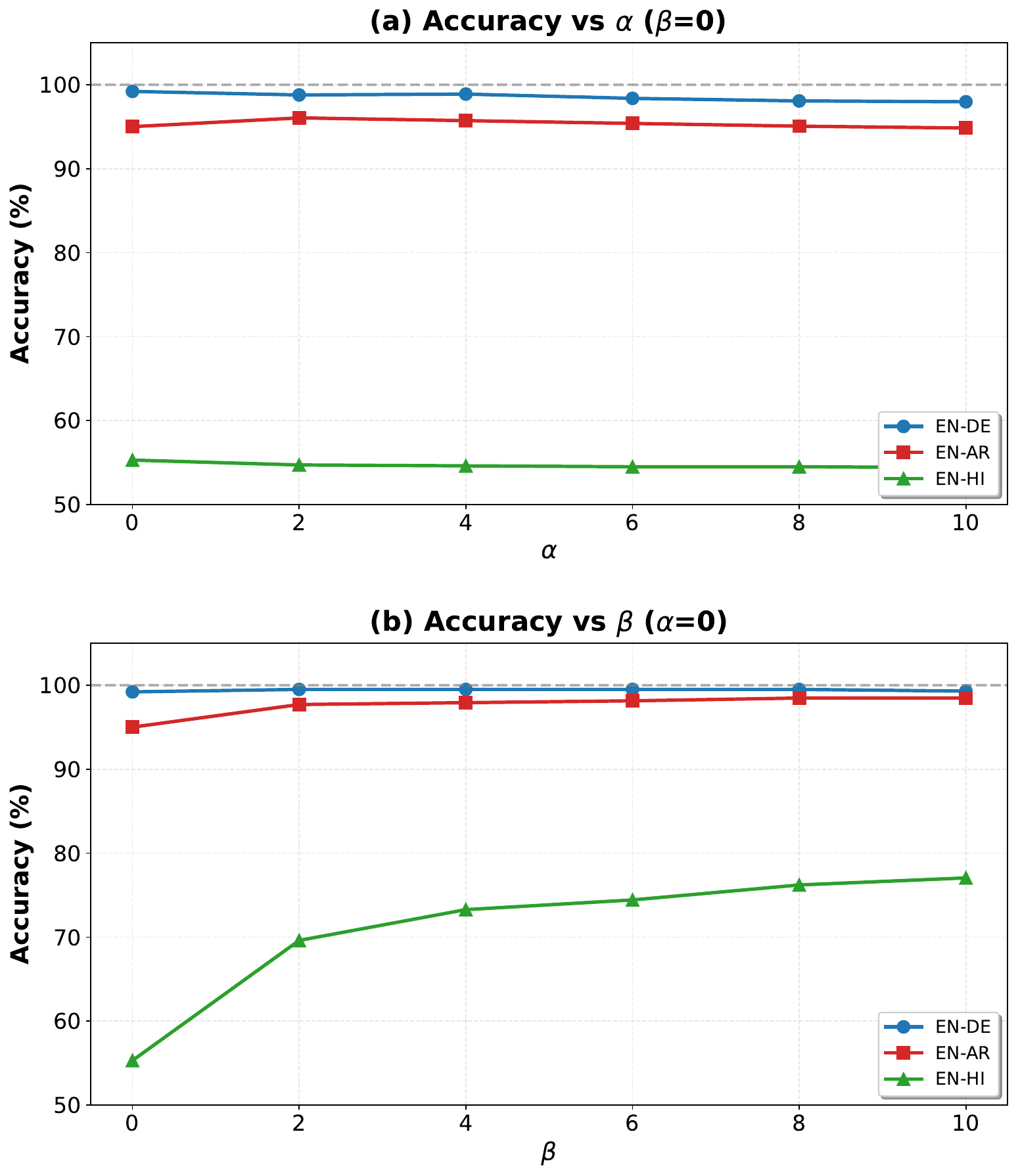}
    \caption{Analysis of hyperparameters $\alpha$ and $\beta$ across three language pairs under gender-explicit setting. Both panels (a) and (b) report binary accuracy, where the gender-aligned translation is scored higher.}
    \label{fig:alpha_beta_exp_vertical}
\end{figure}

As shown in panel (b) of Figure~\ref{fig:alpha_beta_exp_vertical}, the hyperparameter $\beta$ exhibits a larger change in accuracy under gender-explicit settings than under gender-ambiguous settings (see Section \ref{sec:ablation_hyper_params}), indicating a stronger impact in the gender-explicit case.
In contrast, as shown in panel (a), accuracy varies only marginally with respect to $\alpha$, suggesting that the influence of this hyperparameter is reduced.
This behavior aligns with the framework design, as $\beta$ exerts greater influence when explicit gender cues are detected.

\subsection{Analysis with an Additional QE Backbone}
Table~\ref{tab:backbone_results} reports the evaluation results on EN--* language pairs under the gender-ambiguous (Fem. vs. Masc.) setting, where we additionally use COMETKiwi 23 XL to examine the generality of FairQE across different QE backbones. While COMETKiwi 23 XL exhibits larger deviations in gender score ratios compared to COMETKiwi 22, it provides a complementary setting to evaluate the robustness of FairQE under varying bias characteristics.

Overall, FairQE consistently moves the score ratios closer to 1 for most language pairs, with more pronounced improvements in cases where the baseline deviations are larger (e.g., EN--ES, EN--FR, EN--IT).

For EN--DE and EN--HI, where the baseline score ratios are already close to 1, the application of FairQE results in slight overcorrections (i.e., score ratios exceeding 1), indicating a mild shift toward feminine translations. However, for EN--HI, the absolute deviation from 1 is reduced from 0.016 to 0.01, suggesting that the overall fairness is still improved despite the directional shift.

These results demonstrate that the proposed Bias-Aware Score Aggregation is not tied to a specific QE backbone and remains effective across models with different levels of inherent bias. This supports the generality and robustness of the proposed framework.

\subsection{Comparison with Stronger LLM Evaluators}

Table~\ref{tab:strong_llm} reports the evaluation results on EN--* language pairs under the gender-ambiguous (Fem. vs. Masc.) setting, comparing FairQE with direct LLM-based evaluators of different capacities.

While using a stronger LLM (\texttt{o3}) improves fairness over gpt-4.1-mini in most language pairs (except EN--IT), FairQE---despite using \texttt{gpt-4.1-mini}---consistently achieves score ratios closer to 1. In terms of average absolute deviation from 1, \texttt{o3} yields 0.0263, whereas FairQE achieves 0.00638.

Moreover, prior work \cite{zaranis2025watching} shows that even stronger models such as \texttt{gpt-4o} still exhibit greater gender bias under the same setting. These results suggest that fairness improvements are not solely driven by model capacity, but by the structured design of FairQE, which combines contrastive evaluation with bias-aware aggregation.

\subsection{Analysis of Error Propagation from Gender Cue Detection}

To analyze the potential impact of incorrect gender cue detection, we categorize the outputs of $Agent_{cue}$ into four detected cue types: \textit{Gender Ambiguous}, \textit{Gender Explicit}, \textit{Both}, and \textit{None}, and evaluate sentence-level correctness. The analysis is conducted on the EN--DE dataset under two settings: (1) the gender-ambiguous (Fem. vs. Masc.) setting and (2) the gender-explicit setting. In each setting, only one cue type is considered correct, allowing us to assess the extent of misdetection and its downstream effects.

For each detected cue type, we examine both the the QE model’s score ratio (Fem./Masc.) and cue detection accuracy to quantify potential error propagation.

In the gender-ambiguous setting (Table~\ref{tab:cue_ambiguous}), 67.0\% of instances are correctly detected as \textit{Gender Ambiguous}, yielding a near-ideal score ratio of 1.0033. Although 21.6\% of instances are detected as \textit{Gender Explicit}, the corresponding score ratio (0.9925) remains close to 1, indicating minimal degradation in fairness. Instances detected as \textit{Both} or \textit{None} similarly do not exhibit significant fairness deterioration.

In the gender-explicit setting (Table~\ref{tab:cue_explicit}), 91.1\% of instances are correctly detected as \textit{Gender Explicit}, with a detection accuracy of 99.9\%. Instances detected as \textit{Gender Ambiguous} account for only 2.3\% and do not lead to a noticeable drop in fairness performance.

Overall, while cue detection accuracy varies across settings, the impact of misdetection on downstream gender fairness remains limited. These results suggest that error propagation from the $Agent_{cue}$ component is not substantially amplified in the QE evaluation stage.

\subsection{Failure Case Analysis}

Table~\ref{tab:failure_case} presents a representative example from the EN--ES language pair under the gender-ambiguous (Fem. vs. Masc.) setting, where the bias is increased after applying FairQE.

In this example, $Agent_{cue}$ detects a gender cue when the source sentence is paired with the feminine translation, but fails to detect it when paired with the masculine translation. Ideally, the noun \textit{adventurers} should be aligned with its Spanish realizations (\textit{aventureros} / \textit{aventureras}) to enable consistent cue detection. However, masculine plural forms such as \textit{aventureros} are often used as generic masculine expressions in Spanish, referring to mixed-gender groups, which makes them less likely to be detected as explicit gender cues.

Although the $Agent_{uqe}$ score is identical for both variants (95), no cue is detected in the masculine case, resulting in $b_{amb}$ and $b_{exp}$ being set to 0 and leaving the original score $q_{orig}$ unchanged. In contrast, the feminine variant triggers a non-zero $b_{amb}$, leading to an upward adjustment and ultimately reversing the final ranking.

Future work will focus on better understanding how LLMs handle gender-related expressions, as well as systematically analyzing failure patterns and exploring strategies to mitigate them.
\label{sec:B_app}

\section{API Usage and Cost Analysis}
\begin{table}[!ht]
\centering
\small
\renewcommand{\arraystretch}{1.15}
\begin{tabularx}{\columnwidth}{X|r}
\Xhline{3\arrayrulewidth}
\multicolumn{1}{c|}{\textbf{Statistic}} & \multicolumn{1}{c}{\textbf{Amount}} \\
\hline\hline
\multicolumn{2}{c}{\textbf{API Cost (GPT-4.1-mini)}} \\
\hline
Total API calls & 176 \\
Input tokens & 157,192 \\
Output tokens & 19,824 \\
Total tokens & 177,016 \\
Input cost (\$) & 0.0629 \\
Output cost (\$) & 0.0317 \\
\textbf{Total cost (\$)} & \textbf{0.0946} \\
\hline
Avg. tokens / sample & 1,770.2 \\
Avg. cost / sample (\$) & 0.00095 \\
\Xhline{3\arrayrulewidth}
\end{tabularx}
\caption{Estimated API cost statistics scaled to 100 samples using \texttt{GPT-4.1-mini}.}
\label{tab:tab6}
\end{table}

We employ \texttt{GPT-4.1-mini} for all four LLM agents in our framework, and the resulting
API costs are summarized in Table~\ref{tab:tab6}.
Using the EN–DE dataset from the WMT 2023 Metrics Shared Task, we randomly sampled
100 instances and measured the corresponding API usage.
The total number of API calls amounts to 176, which is substantially lower than
the maximum of 400 calls required when all four agents are invoked for every sample.
This reduction is achieved by a dynamic and efficient invocation strategy, where
subsequent agents are selectively triggered based on the outputs of the gender cue
detection agent.
Overall, the total cost per 100 samples is 0.0946 USD, indicating that the proposed
framework can be executed with a cost of less than 0.1 USD per 100 samples.

\label{sec:C_app}

\section{AI Assistant Usage}
We have used Claude Code during the development of our research work.
\label{sec:D_app}

\clearpage
\onecolumn

\section{Gender Cue Taxonomy}
\begin{table}[!ht]
\centering
\small
\renewcommand{\arraystretch}{1.2}
\begin{tabularx}{\columnwidth}{c|c|X|X}
\Xhline{3\arrayrulewidth}
\textbf{Cue} & \textbf{Type} & \textbf{Description} & \textbf{Examples} \\
\hline\hline
C1 & Explicit &
Gendered pronouns that directly indicate gender. &
he / she, him / her, his / her \\

C2 & Explicit &
Gender-fixed kinship nouns inherently tied to gender. &
mother, father, sister, brother, uncle, aunt \\

C3 & Explicit &
Gendered noun pairs with lexical gender distinction. &
actor / actress, waiter / waitress \\

C4 & Explicit &
Titles or honorifics with explicit gender marking. &
Mr., Ms., Mrs., señor / señora \\

C5 & Explicit &
Speaker-gender-marking expressions that encode the speaker’s gender directly. &
Japanese \begin{CJK}{UTF8}{min}\textsf{僕 / 私}\end{CJK}; Arabic gender-marked verbs \\

C6 & Explicit &
Gender agreement requirements where morphological forms must match gender. &
noun–adjective agreement in Romance languages \\

\hline
C7 & Ambiguous &
Gender-neutral occupation or role nouns without gender information in the source. &
doctor, teacher, engineer \\

C8 & Ambiguous &
Gender-neutral pronouns or indefinites. &
singular they / them / their; someone, anybody \\

C9 & Ambiguous &
Gender-unknown proper names where gender cannot be reliably inferred. &
Alex, Sam \\

C10 & Ambiguous &
Subject omission or passive constructions where agent gender is unspecified. &
``Arrived early'', agentless passive \\

C11 & Ambiguous &
Neutral relation nouns that do not encode gender. &
colleague, partner, friend \\

C12 & Ambiguous &
Generic or plural group references without gender specification. &
doctors, people, students \\
\Xhline{3\arrayrulewidth}
\end{tabularx}
\caption{Cue taxonomy for gender-related signals, grouped into explicit (C1--C6) and ambiguous (C7--C12) categories.}
\label{tab:tab7}
\end{table}

\clearpage
\label{sec:E_app}

\section{Prompt Construction for FairQE}
\begin{tcolorbox}[width=\textwidth, colback=gray!5, colframe=black, boxrule=0.5pt, title=Gender Cue Detector ($Agent_{\textit{cue}}$)]
\small
\ttfamily
\obeylines
\obeyspaces
\textbf{SYSTEM\_PROMPT} = """You are a Gender Cue Detection Agent.
~
Your ONLY job:
  - Detect gender-related cues in BOTH source and target sentences.
~
Decision procedure:
  1) Examine the source sentence only.
  2) If the source contains any explicit gender marker (C1--C6), classify as \hspace*{1em} gender\_explicit.
  3) Otherwise, if the source contains gender-neutral expressions or lacks \hspace*{1em} gender information, classify as gender\_ambiguous.
  4) Use the target sentence only to align corresponding expressions, not to \hspace*{1em} determine ambiguity or explicitness.
~
Hard constraints:
  - Do NOT judge translation quality.
  - If no gender-related cues (C1--C12) are found in BOTH source and target, \hspace*{1em} return an empty JSON object \{\}.
  - Output JSON only.
~
Output schema (JSON object only):
  \{
    "gender\_ambiguous": [
      \{"source\_token": string|null, "target\_token": string|null\}, ...],
    "gender\_explicit": [
      \{"source\_token": string|null, "target\_token": string|null\}, ...]
  \}
~
Cue taxonomy (C1--C12):
~
[Explicit cues: C1--C6]
C1 (Explicit) Gendered pronouns:
  - Pronouns that directly indicate gender
\hspace*{1em}(he/she, him/her, his/her, etc.)
~
C2 (Explicit) Gender-fixed kinship nouns:
  - Kinship terms inherently tied to gender
    \hspace*{1em}(mother, father, sister, brother, uncle, aunt, etc.)
~
C3 (Explicit) Gendered noun pairs:
  - Lexical pairs with gender distinction
    \hspace*{1em}(actor/actress, waiter/waitress, etc.)
~
C4 (Explicit) Titles / honorifics:
  - Explicit gender markers in titles or honorifics
    \hspace*{1em}(Mr., Ms., Mrs., señor/señora, etc.)
~
C5 (Explicit) Speaker-gender-marking expressions:
- Source/target forms that encode speaker gender (e.g., Japanese \jp{僕 / 私};  Arabic \hspace*{1em} gender-marked verbs, etc.)
- Note: Detect as explicit when the expression itself marks speaker gender.
~
C6 (Explicit) Gender agreement requirements:
  - Morphological agreement that must match gender
    \hspace*{1em}(e.g., pronoun--adjective or noun--adjective endings in Romance languages)
  - Detect mismatches as cues present (NOT as errors);only record where such \hspace*{1em}agreement markers appear.

\end{tcolorbox}

\clearpage

\begin{tcolorbox}[width=\textwidth, colback=gray!5, colframe=black, boxrule=0.5pt, notitle]
\small
\ttfamily
\obeylines
\obeyspaces

[Ambiguous cues: C7--C12]
C7 (Ambiguous) Gender-neutral occupation or role nouns:
  - Role or occupation nouns without gender information in the source (doctor, \hspace*{1em} teacher, engineer, etc.)
~  
C8 (Ambiguous) Gender-neutral pronouns / indefinites:
  - they/them/their (singular), someone, anybody, etc.
~
C9 (Ambiguous) Gender-unknown proper names:
  - Names where gender cannot be reliably inferred
    \hspace*{1em}(Alex, Sam, etc.)
~
C10 (Ambiguous) Subject omission / passive constructions:
  - Source does not specify agent gender
    \hspace*{1em}(e.g., ``Arrived early'', agentless passive voice)
~
C11 (Ambiguous) Neutral relation nouns:
  - colleague, partner, friend, etc.
~
C12 (Ambiguous) Generic group or generalization:
  - plural or generic references without specifying gender (doctors, people, \hspace*{1em} students, etc.)
"""
~
\textbf{USER\_PROMPT} = """
Source: ```\textcolor{blue}{\{source\}}```
Target: ```\textcolor{blue}{\{target\}}```
"""
\end{tcolorbox}


\begin{tcolorbox}[width=\textwidth, colback=gray!5, colframe=black, boxrule=0.5pt, title=Gender-Ambiguous Variant Generator ($Agent_{\textit{amb}}$)]
\small
\ttfamily
\obeylines
\obeyspaces
\textbf{SYSTEM\_PROMPT} = """You are a Gender-Ambiguous Variant Generator.
~
You ONLY handle cases where the Gender Cue Detection Agent has identified gender\_ambiguous cues in the source sentence.
~
Your job:
  - Generate alternative gender versions of the target sentence by WORD-LEVEL \hspace*{1em} substitution only,
  - Using ONLY the target\_token positions provided by the Gender Cue Detection \hspace*{1em} Agent as anchors.
~
Hard constraints:
  - NO paraphrase and NO sentence restructuring. Keep punctuation, word order, \hspace*{1em} and all other tokens unchanged.
  - ONLY substitute gender-related words or phrases that correspond to the Gender \hspace*{1em} Cue Detection Agent's ambiguous cues.
  - If substitution is impossible, return an empty list [].
  - Generate only linguistically natural and contextually valid versions.
~
Output schema (JSON object only):
[
  \{"transformed\_text": string,
    "gender\_version": "Feminine" | "Masculine" | "Neutral"\},
  ...
]
"""
~
\textbf{USER\_PROMPT} = """
Source: ```\textcolor{blue}{\{source\}}```
Target: ```\textcolor{blue}{\{target\}}```

Gender Cue Detection Agent's ambiguous cues: \textcolor{blue}{\{ambiguous\_pairs\_json\}}
"""
\end{tcolorbox}


\begin{tcolorbox}[width=\textwidth, colback=gray!5, colframe=black, boxrule=0.5pt, title=Gender-Explicit Variant Generator ($Agent_{\textit{exp}}$)]
\small
\ttfamily
\obeylines
\obeyspaces
\textbf{SYSTEM\_PROMPT} = """You are a Gender-Explicit Variant Generator.
~
You ONLY handle cases where the Gender Cue Detection Agent has identified gender\_explicit cues in the source sentence.
~
Your job:
  1) Using ONLY the explicit gender cues provided by the Gender Cue Detection \hspace*{1em} Agent as anchors, compare the source and target sentences to verify whether \hspace*{1em} explicit gender constraints are preserved.
  2) Detect the following violations:
     - gender flip (e.g., feminine $\rightarrow$ masculine or vice versa),
     - gender agreement errors (e.g., pronouns or gendered nouns),
     - clear mismatches for gender-fixed expressions.
  3) Set error = True ONLY if such violations exist.
~
Decision logic:
  - If error == True:
      \hspace*{1em}Generate corrected versions of the target sentence by WORD-LEVEL substitution ONLY.
  - If error == False:
      \hspace*{1em}Generate gender-flipped versions of the target sentence by WORD-LEVEL \hspace*{1em} substitution ONLY.
~
Hard constraints:
  - NO paraphrase and NO sentence restructuring. Keep punctuation, word order, \hspace*{1em} and all other tokens unchanged.
  - ONLY substitute gender-related words or phrases that correspond to the Gender \hspace*{1em} Cue Detection Agent's explicit cues.
  - If substitution is impossible, return an empty list [].
  - Generate only linguistically natural and contextually valid versions.
~
Output schema (JSON object only):
[
  {
    "error": boolean,
    "rationale": string,
    "transformed": [
      \{"transformed\_text": string, "gender\_version": "Feminine" | "Masculine" | "Neutral"\},
      ...
    ]
  }
]
"""
~
\textbf{USER\_PROMPT} = """
Source: ```\textcolor{blue}{\{source\}}```
Target: ```\textcolor{blue}{\{target\}}```

Gender Cue Detection Agent's explicit cues: \textcolor{blue}{\{explicit\_pairs\_json\}}
"""
\end{tcolorbox}


\begin{tcolorbox}[width=\textwidth, colback=gray!5, colframe=black, boxrule=0.5pt, title= Bias-Mitigating Quality Estimator ($Agent_{\textit{uqe}}$)]
\small
\ttfamily
\obeylines
\obeyspaces

\textbf{SYSTEM\_PROMPT} = """You are an Unbiased QE Scorer.
~
Your task is to evaluate translation quality using an MQM-style protocol,
while remaining as gender-independent as possible by following the rules below.
~
MQM Evaluation Rules:
  - Based on the source segment and the machine translation enclosed in triple \hspace*{1em} backticks, identify and classify ALL translation errors.
  - Error types include:
    \hspace*{1em}* accuracy (addition, mistranslation, omission, untranslated text)
    \hspace*{1em}* fluency (character encoding, grammar, inconsistency, punctuation, register, \hspace*{1em} spelling)
    \hspace*{1em}* locale convention (currency, date, name, telephone, time format)
    \hspace*{1em}* style (awkward)
    \hspace*{1em}* terminology (inappropriate for context, inconsistent use)
    \hspace*{1em}* non-translation
    \hspace*{1em}* other
    \hspace*{1em}* or no-error
  - For EACH identified error, assign a severity level:
    \hspace*{1em}* Critical
    \hspace*{1em}* Major
    \hspace*{1em}* Minor
~
Scoring:
  - Start from a score of 100 points.
  - Deduct points as follows:
    \hspace*{1em}* Critical: --15 points
    \hspace*{1em}* Major: --5 points
    \hspace*{1em}* Minor: --1 point
  - The final score must be between 0 and 100.
  - Optionally, apply a holistic Direct Assessment (DA)-style judgment ONLY \hspace*{1em} if the MQM-based score clearly under- or over-estimates overall translation \hspace*{1em} quality.
~
You are given:
  - a source sentence,
  - its original translation,
  - and gender-flipped target translations generated by substituting ONLY \hspace*{1em} gender-related expressions.
~
Use the gender-flipped translations to compare them against the source and the original translation, and determine whether meaning is preserved.
~
1) Gender-Ambiguous Source Cases
  - The source contains no explicit gender information.
  - Gender-flipped translations differ ONLY in gender expression and are all \hspace*{1em} valid (Feminine / Masculine / Neutral).
~
Rules:
  - Gender differences MUST NOT affect the quality score.
  - The Neutral form MAY be preferred if it is most natural, but this preference \hspace*{1em} MUST NOT lower the scores of Feminine or Masculine variants.
~
2) Gender-Explicit Source Cases
  - The source specifies a clear gender constraint.
  - A gender error flag (error) and its explanation are provided.
~
  - If error == True:
      \hspace*{1em} Gender-corrected translations are provided and MUST be reflected as MQM \hspace*{1em} errors with appropriate severity.
  - If error == False:
      \hspace*{1em} Alternative gender variants (0--2 among Feminine / Masculine / Neutral) are \hspace*{1em} provided and used to assess whether the original translation deserves an \hspace*{1em} appropriate score.
~
Rules:
  - Violations of explicit gender constraints MUST be marked as MQM errors with \hspace*{1em} appropriate severity.

\end{tcolorbox}

\clearpage

\begin{tcolorbox}[width=\textwidth, colback=gray!5, colframe=black, boxrule=0.5pt, notitle]
\small
\ttfamily
\obeylines
\obeyspaces

Output schema (JSON object only):
\{
  "qe\_score": number,
  "rationale": string
\}
"""
~
\textbf{USER\_PROMPT} = """
Source: ```\textcolor{blue}{\{source\}}```
Target: ```\textcolor{blue}{\{target\}}```

Gender cues:
  - ambiguous: \textcolor{blue}{\{ambiguous\_pairs\_json\}}
  - explicit: \textcolor{blue}{\{explicit\_pairs\_json\}}

Gender-Ambiguous source cases
(with gender-flipped target translations): \textcolor{blue}{\{amb\_alternatives\_text\}}

Gender-Explicit source cases
(with error analysis and gender-flipped target translations): \textcolor{blue}{\{exp\_analysis\_text\}}
"""
\end{tcolorbox}

\end{document}